# Hybrid CNN-LSTM based Indoor Pedestrian Localization with CSI Fingerprint Maps

**Muhammad Emad-ud-din**

Department of Computer Science and Engineering

Texas A&M University, College Station, TX

emaad22@tamu.edu

## Abstract

*The paper presents a novel Wi-Fi fingerprinting system that uses Channel State Information (CSI) data for fine-grained pedestrian localization. The proposed system exploits the frequency diversity and spatial diversity of the features extracted from CSI data to generate a 2D+channel image termed as a "CSI Fingerprint Map". We then use this CSI Fingerprint Map representation of CSI data to generate a pedestrian trajectory hypothesis using a hybrid architecture that combines a Convolutional Neural Network and a Long Short-Term Memory Recurrent Neural Network model. The proposed architecture exploits the temporal and spatial relationship information among the CSI data observations gathered at neighboring locations. A particle filter is then employed to separate out the most likely hypothesis matching a human walk model. The experimental performance of our method is compared to existing deep learning localization methods such ConFi, DeepFi and to a self-developed temporal-feature based LSTM based location classifier. The experimental results show marked improvement with an average RMSE of 0.36 m in a moderately dynamic and 0.17 m in a static environment. Our method is essentially a proof of concept that with (1) sparse availability of observations, (2) limited infrastructure requirements, (3) moderate level of short-term and long-term noise in the training and testing environment, reliable fine-grained Wi-Fi based pedestrian localization is a potential option.*

## 1. Introduction

People localization is an essential component of many applications like indoor security camera systems, activity classification, and elderly people who need to be monitored. Apart from Wi-Fi based localization, alternate options like camera-based people tracking generally fail to deliver because of problems with maintaining line-of-sight with the sensor, privacy concerns, variance in lightening conditions, high velocity motions, computational and infrastructure costs. Apart from cameras, sensors like IMUs and Magnetometers are either too noisy or expensive to be part of a scalable solution.

Since Wi-Fi infrastructure is ubiquitous in indoor environments, many have jumped on the opportunity and designed efficient and scalable localization systems based on deep learning models. These systems rely on either active or passive localization methods. Passive methods include scenarios where users carry no Wi-Fi device and are localized based on the available Wi-Fi signals that are reflected from their bodies [11-14]. Active methods primarily rely on user carried Wi-Fi devices for gathering fingerprints usually extracted from either Received Signal Strength Indicator (RSSI) measure or Channel State Information (CSI). A cost-effective and scalable solution which has a potential to provide localization without the need of any additional signal or infrastructure, seems like a promising road to a future widespread adoption for this technology.

Our approach towards pedestrian localization involves an active method of localization that pivots on two assumptions (1) The CSI measure, apart from being a tool to monitor channel conditions for achieving high data-rates, can also be used as a spatially diverse signal. (2) The fingerprinting strategy, the pre-processing and the post-processing framework used to shape the training data associated to indoor locations, ensures that fingerprint features remain diverse over space and robust to noise over longer durations of time. As such, there exists wide support in the literature for the fact that CSI can capture fine-grained variations over space in the wireless channel, but no formal study has been conducted to study the signal stability over time.

Given the above pre-requisites, we pre-processed the CSI observations so that these remain temporally stable for pedestrian positioning task over extended periods of time. Spatial diversity in the CSI measure exists by the virtue of a large number of signal features it considers but it is certainly a challenge to extract a temporally stable fingerprint formulation considering how noisy these signals can get. This happens due to the extent of Radio Frequency (RF) phenomena present in an uncontrolled environment. Figure 1 shows two smoothed CSI Fingerprint Maps for the same channel over a 6m by 4m space, side-by-side, taken at different points in time. This figure highlights the high level of change in the signal over a period of 44 days in the spatial distribution of CSI data. This change in the carrier frequency response over time brings in temporal instability to the features that we are aiming to use as fingerprints for localization. *Fingerprints* in our context, can be referred to as a temporally stable set of data which can be uniquely associated to a location in space. We propose a set of de-noising measures in Section 3, that directly address the temporal instability in the signal.

In brief, we investigate the use of a 2D+channels representation of CSI data to train a hybrid architecture that consists of a CNN (Convolutional Neural Network) and LSTM (Long Short-Term Memory) network, for achieving pedestrian positioning. Our work makes the following key contributions. (1) We realize a deep learning framework that can achieve superior and robust pedestrian localization, via a novel fingerprint representation called *CSI Fingerprint Map*. (2) Robust localization is achieved in the presence of both short-term and long-term signal noise. (3) A further improvement in pedestrian positioning and heading estimate accuracy is achieved via a particle filter based post-processing step. This step filters out mis-classified pedestrian positions via analyzing multiple hypothesis of pedestrian trajectories. We thus propose a localization solution which is an end-to-end pipeline that begins with CSI data collection, transitions into fingerprint formulation, then performs training and learning of a model, and finally filters the model output via a particle filter for submitted query fingerprints. Our method trains a CNN based on a set of sliding window sizes. In this context, we introduce a novel notion of *Information Adaptive Proposal Size* which is how we adjust the size of sliding window to generate training proposals for our CNN model. The intuition here is that during the test phase, as the pedestrian explores the environment, the spatial diversity of the test fingerprint can be increased by including more information. We will explain this in detail section 4.

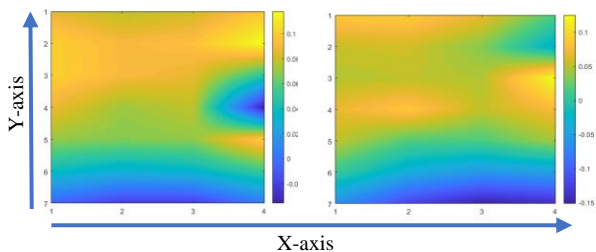

Figure 1. Difference in CSI data captured for the same sub-carrier frequency index. Data is plotted across area of testbed (Left) data captured on 15 Feb 2019 (Right) data captured on 9 April 2019. Blue represents low values for sanitized phase component of CSI measure while yellow represents high values.

After presenting literature review in section 2, CSI metric details and signal denoising are explained in section 3. Details about the proposed method and the dataset are given in section 4. In section 5, we show the effect of method parameters on localization accuracy. These parameters include map grid-size for training dataset collection, CNN sliding window size, accuracy before and after the application of particle filter based post-processing step. We also evaluate the performance of the proposed method and compare it to closely related methods by utilizing a self-collected long-term dataset in the same section. Section 6 lists down the summary and conclusion of our work.

## 2. Literature Review

For coarse-grained localization, RSSI based fingerprinting has not been able to show convincing accuracy, primarily because of lack of space diversity in RSSI measure. Both RSSI and CSI channel quality indicators contain unpredictable variance over time and space due to the presence of people and existing obstacles, which create RF phenomena like reflection, refraction and multipath interference, thus impairing precise positioning. Even in the case of CSI, very close locations usually share similar fingerprints and, therefore, the typical average accuracy of Wi-Fi positioning systems is within a few meters (~1–4m) [5]. In an effort to achieve fine-grained localization, some complementary approaches have been suggested that make use of a range of noisy sensors like IMUs (Inertial Measurement Units) [6], light intensity sensors and magnetometers [7] but these suffer from sources of error like nonlinear and time-dependent noise and accumulative error [8]. Thus, traditional methods use reliable external sources to reset the bias and accumulative error. A more recent work [4] affirms the overall superiority of CSI over RSSI in terms of stability for effective localization via fingerprinting. It also highlights that the difference of CSI phase values between two antennas for 5GHz orthogonal frequency-division multiplexing (OFDM) channel, is highly stable at a fixed location compared to RSSI channel information or similar 2.4GHz OFDM channel CSI phase values. Such revelations have dictated our choice of 5 GHz OFDM channel for measuring CSI phase difference values between two antennae on the same device. A higher-level derived feature (computed from calibrated phase difference between antenna), such as angle of arriving (AOA) could also be used but requires that there is a strong Line of Sight (LOS) component between the access point (AP) and the receiver (which may not be the case for our collected dataset). The simulation work presented in [5] claims that since real RSSI datasets would always contain significant noise due to environmental factors, working with simulated RSSI signals has obvious advantages since we have a ground-truth signal available. Authors in [5] have shown that they have the liberty to introduce controlled noise and error-sources and study its impact on fingerprints in isolation to the noise introduced by uncontrolled factors. This study is particularly useful for determining the impact of distance and presence of other RF signal sources on the spatial diversity of signal.

We also evaluated popular RSSI datasets available at [10] for long term fingerprint degeneration. It was noticed that an average of signal strength difference of up to a maximum of 17 dBm was present observed over a period of 6 months at a university in Tampere, Finland [9]. These differences are bound to translate into localization accuracy degradation over time, but as of now no relevant studies are available that quantify such a degradation. Moreover, no obvious relationship was observed between signal strength and observed variance in RSSI values over time. In other words, high variance in RSSI signal was observed at almost all locations regardless of the signal strength. It may be noted here that due to the lack of comprehensive CSI based datasets and scarcity of studies using these datasets to comment on spatial and temporal characteristics of CSI metric, we resorted to extrapolation and deduction with respect to CSI metric behavior in many cases from RSSI based studies.

We also looked at the recent trends for using deep learning for Wi-Fi based localization systems, more specifically, CSI fingerprinting based positioning systems. A CSI fingerprinting based system known as DeepFi was proposed in [15] with four layers neural network. DeepFi was able to improve positioning accuracy by at least 20% compared to the best available accuracy for an RSSI based localization system. A system called CiFi, proposed in [16], uses a convolutional network (CNN) for indoor localization based on Wi-Fi signals. Here the CSI phase data is used to estimate the angle of arrival (AOA) which is simply reshaped into a single image data and sent as an input to the convolutional network. The results show that CiFi has an error of less than 1 m for 40% of the test locations which is far superior than previously mentioned methods. Another system, termed ConFi was proposed [17], which is a CNN based Wi-Fi localization technique that uses CSI as features. To be more specific, the CSI was organized as a CSI feature image, with observation time at one axis, subcarrier frequency amplitudes at the other axis and frequency amplitude for each sending and receiving antenna pair at one axis. The network is trained using these CSI feature images. ConFi reduced the mean positioning error by 9.2% over DeepFi. Another method [19] that uses 1D CNN for localization while using a raw Radiofrequency (RF) feature called I/Q imbalance, claims an accuracy of 90-99% accuracy. Both CSI Image representations mentioned above fail to relate the 2D expanse of an image representation to the 2D expanse of a physical map locations. The proposed relationship in this respect is explained in detail in Figure 2.

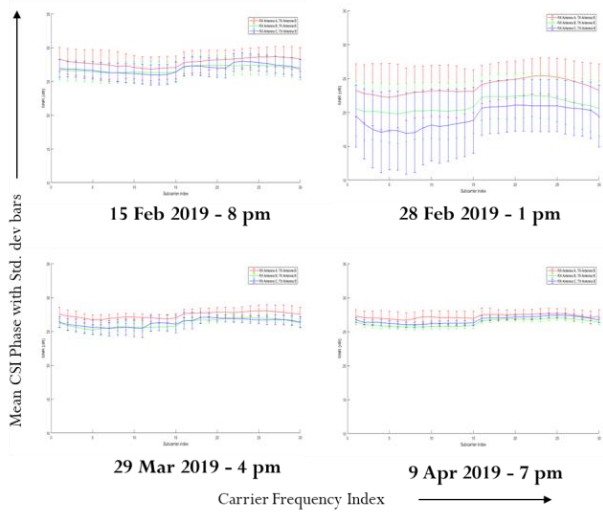

Figure 2. Mean CSI Phase values for 30 subcarrier frequencies along with associated standard deviation bound. Data shown for location (0,6), TX=B and RX=A, B, C for 4 different observations gathered during different times over a period of two months.

As per all the literature surveyed, no instance was found where RSSI or CSI data was analyzed for spatial and long-term temporal diversity for estimation using a map-based re-arrangement of CSI features for position estimation.

## 3. Preliminaries and CSI Phase Information

### 3.1. Channel State Information

CSI records the variation in an 802.11n standard channel, experienced during propagation. A wireless signal may experience variance caused due to possible RF phenomena like the multipath effect, fading, shadowing, reflection, refraction, interference and delay distortion. Without a channel quality metric like CSI, it is almost impossible to analyze the channel characteristics with only the signal power. Briefly, CSI represents the channel's frequency response, which can be estimated from transmitted and received signal vectors.

The 5GHz band Wi-Fi channel is a narrowband flat fading channel. We use Intel Wi-Fi Link 5300 NIC wireless LAN card that can read 30 subcarriers out of a total 56 subcarriers for CSI information via the device driver. The channel frequency response $CSI_i$ of subcarrier $i$ is a complex value, which can be represented as follows

$$CSI_i = |CSI_i|e^{jsin(\angle CSI_i)} \quad (1)$$

where $|CSI_i|$ and $\angle CSI_i$ are the amplitude response and the phase response of subcarrier $i$, respectively. Our fingerprinting approach uses sanitized phase values for 30 subcarriers in the OFDM system, which when used with a multiple-input and multiple-output (MIMO) router and a wireless receiver device both having 3 antennae, translates into a total of 270 phase values.

### 3.2. Sanitizing CSI phase values

Though the CSI phase information is readily available from the Intel 5300 NIC, it cannot be directly used for localization, due to noise and the unsynchronized time and frequency of the transmitter and receiver. We use a linear transform-based approach listed in [18] to sanitize the phase values measured from the 30 subcarriers. The key idea behind removing the randomness in CSI phase is that the true phase formulation contains certain unknown values i.e. timing offset at the receiver and phase offset, which can be eliminated by considering phase across the entire frequency band. Figure 3 illustrates the phase for location (0,0) from our collected dataset, after transformation, which depicts relatively stable features over a duration of two months compared to the random version (not shown). We do not claim that sanitized information is equal to the true phase but for purposes of fingerprinting diversity and stability, sanitized phase in its current form is a usable and effective feature.

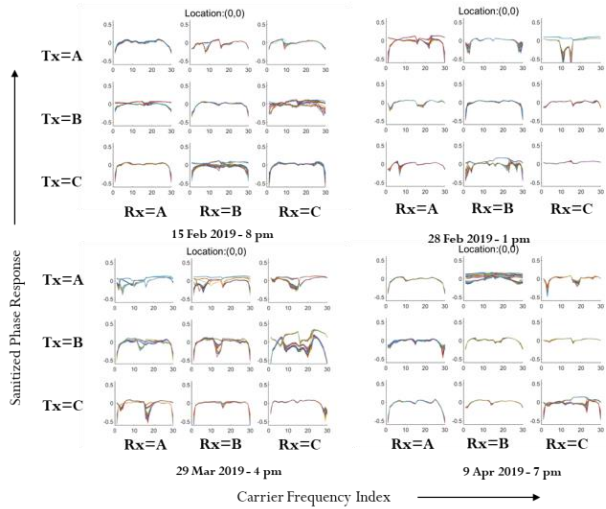

Figure 3. Stability of sanitized, normalized, phase values for location (0,0) over a two-month period

### 3.3. Denoising the CSI values

We performed a detailed time-based analysis of variance present in CSI amplitude response and phase response. The analysis is based on a dataset gathered over a period of almost two months. This analysis precipitated the following conclusions which helped shape our decision as to what CSI component to choose for fingerprinting and decide a denoising approach to remove variance caused due to RF phenomenon caused by the environment.

- For a location, the CSI values for both amplitude and phase for different antennas pairs (transmitting and receiving antenna pairs) show unique patterns. Thus, using CSI values corresponding to more antenna pairs is bound to bring more diversity to our fingerprint.

- For a location, CSI values on certain subcarriers frequencies occasionally show very high variance due to RF propagation phenomena discussed earlier. Skipping these frequencies altogether from the dataset does not justify the accuracy loss as certain "discriminant features" are lost in the process.

- For a given location, for consecutive observations, not separated in time by more than 200 milliseconds, CSI values have very low variance over a short time window e.g. 2 seconds. Under rare environmental circumstances, this variance however grows by a large margin due to RF phenomenon caused by environmental changes such as very frequent motion in the vicinity of access point or Wi-Fi device.

Last observation proved to be essential as literature [5] suggests that 68.27%,95.45% and 99.73% of noisy RSSI values fall within the first, second and third standard deviation ($\sigma$) bounds respectively. We already know that assuming the noise distribution in wireless signals to be a Gaussian is an oversimplification, but this assumption significantly simplifies the computations with little performance loss [5]. We thus assume similar noise distribution and bounds for CSI phase values and remove noisy observations which contain CSI phase values beyond $2\sigma$ bound for any subcarrier frequency. Figure 3 shows the mean CSI phase values gathered over a period of two months, for a location (0,6) in our dataset, along with corresponding standard deviation bounds for each CSI sub-carrier index.

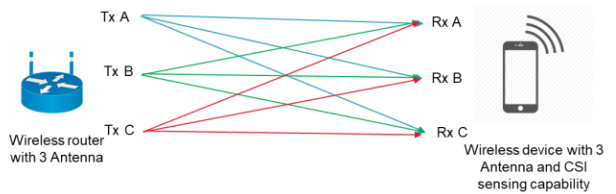

Figure 4. A three transmitter and three receiver antenna MIMO configuration used for dataset collection

## 4. Hybrid CNN LSTM based Pedestrian Localization

### 4.1. Dataset Collection

A CSI observation dataset was collected in the corridors of Emerging Technologies Building at Texas A&M University. Each training observation consists of 30 sanitized CSI Phase values for 9 MIMO channel pairs between an access point (AP) and a Wi-Fi device as shown in Figure 4. Thus, each fingerprint observation collected at location $(x, y)$ can be represented by a set $FP_{(x,y)}^t$

$$FP_{x,y}^t = \{pr_1^t, pr_2^t, \ldots, pr_{270}^t\} \qquad (2)$$

Here $pr_1^t$ represents a sanitized phase value collected at location $(x, y)$ at time $t$. A comprehensive snapshot of the dataset attributes is presented in table 1.

A total of 14400 observations distributed among 36 locations were gathered during the data collection exercise. Each of the observations was time-stamped using suitable resolution so that observations can be used in time-sensitive recurrent networks. More specifically, the following conditions were met to establish the suitability of this dataset to serve as fingerprinting database for a Wi-Fi based positioning system.

a. The training data locations were kept at fixed distance

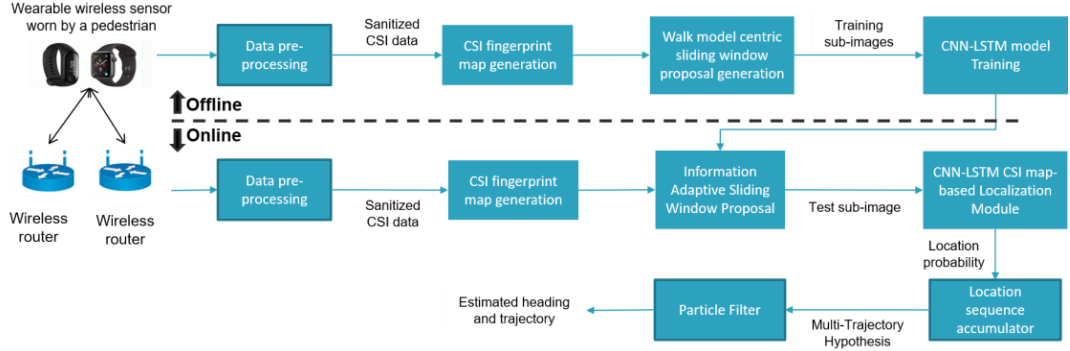

Figure 5. Flowchart for the proposed hybrid CNN-LSTM localization system.

from each other i.e. 1 meter apart.

Table 1

| Attributes | ETB Corridor Dataset |
|---|---|
| Time stamps accuracy | milliseconds |
| Spatial Resolution Unit | meters |
| Data Collection Resolution | 1m |
| Number of classes (locations) | 36 |
| RSSI Features per observation | 6 |
| CSI Features per observation | 270 |
| Observations per location | 400 |
| Total observations | 14400 |
| Temporal Distribution of observations | min: 200 ms apart max: 44 days apart |
| Area | Indoor area: 140 sq. m |
| # of Devices, # of Access points | 1 collection device and 1 access point |

b. The test data was collected at random locations but in a sequential manner and at most 200 milliseconds apart in time, within the 140 sq. m test-bed area. The test dataset was collected at both off-peak and peak hours in terms of pedestrian traffic at the testbed. During peak hours, people activity in the area caused visible variance in CSI data. After denoising the test dataset, we apply the bounds on $t$ in $FP_{x,y}$ so that only a small window of temporal observations can be considered for test fingerprinting. The time window to generate test fingerprints was kept small to limit the amount of variance in our test CSI fingerprints.

c. Apart from collecting CSI Phase values, corresponding CSI Amplitude and RSSI values for each channel pair were also collected for future use and comparative analysis.

**4.2. Proposed Method**

The proposed method and its components are shown in a flowchart illustrated in figure 5. It can be seen in the figure, that our method contains two kinds of dataflows. One is the offline data flow that is meant to collect the CSI fingerprints at certain locations and then train a deep learning model which can be queried for a location when provided with a test fingerprint. The second data flow obviously is the online or the test dataflow. Its job is to collect test CSI fingerprints and feed it to the trained deep learning model to elicit a location estimation. Now going into further detail, the step-by-step method is listed below.

*4.2.1. CSI Fingerprint map*

The RF fingerprint map shown in figure 6, can establish a correspondence between our single observation which is essentially 270 phase responses extracted from Channel State Information (CSI) at each reference location, to a pixel in an image (or a map). This is done via creating a 2D+270 channel representation. The X-Y plane in figure 7 represents the spatial expanse of our testbed while the Z-axis (CSI phase) represents the features or the observation domain for our representation. It may be highlighted here that a single observation collected at time $t$ at a given location can only add a fingerprint $FP_{x,y}^t$ to the image pixel $(x,y)$. To be able to complete a map where all locations have observations, one must gather observations at each location at different times during the dataset collection or test phase. We now have an intuitive representation of CSI data collected over several points in time, arranged in form of an image. We are now ready to exploit this representation by using a sliding window-based method similar to the popular Regional-Convolutional Neural Network(R-CNN). Let us first denote an expression for our CSI Fingerprint map $M^i$.

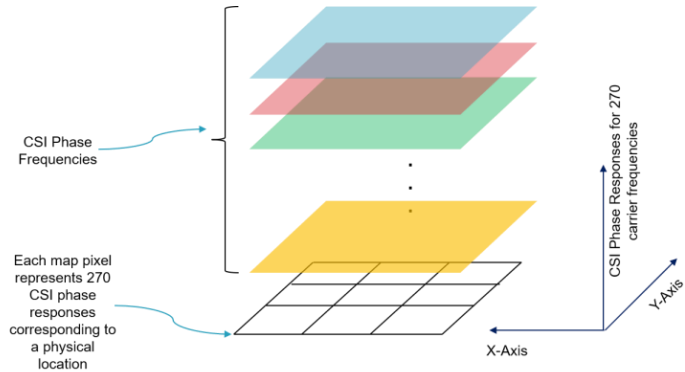

Figure 6. 2D + channels CSI Fingerprint map representation. Here X & Y axis represent the X and Y coordinates of the test-bed area. The vertical axis represents the 270 CSI phase responses collected during a single observation. This each pixel of our CSI Fingerprint Map has 270 channels. 14400 CSI fingerprint maps were collected over a period of 44 days.

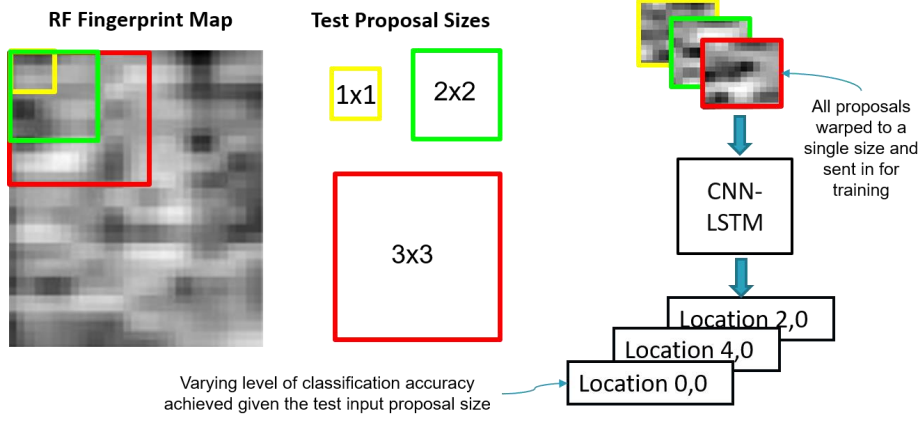

Figure 7. Information Adaptive Sliding Window Proposal scheme

$$M_{x,y}^i = FP_{x,y}^i \qquad (3)$$

Here map observation $M^i$ is a collection of fingerprints $FP_{x,y}^i$, each gathered at distinct reference points on the map (training locations that are 1 meter apart). $i$ here represents the observation index at each location. Notably for each $M^i$, $t'$ has a constant value between 1 to $m$. Where, $m$ is the constant number of fingerprint observations gathered at each reference point on the map. One interesting property of $M$ is that all pixel observations $FP_{x,y}^i \in M^i$ are taken at the previous time step when compared to $FP_{x,y}^{i+1} \in M^{i+1}$. The reason behind why this order is maintained between subsequent observations, $M^i$, is that this enables us to feed these time-ordered maps to a sequential neural network such as LSTM. Thus, this property of "*time-ordered maps*" is key enabler that lets us use these observations in a Hybrid CNN-LSTM model. We also extend this property to the *time-ordered datasets* where each dataset is a set containing a set of time-ordered map observations $\{M^1, M^2, M^3, \dots\}$.

*4.2.2. Sliding Window based Localization Method*

A graphical illustration of this method is presented in figure 7. We list down the proposed method in brief steps listed below.

a. We shape the collected dataset into a 2D+channels CSI Map format. This map $M^i$ has each pixel associated to a location $(x, y)$ where training observations were gathered during the pre-requisite fingerprinting collection stage.

This step explains the notion of *Information Adaptive Sliding Window (Proposal) Size*. We assume a motion model of a pedestrian where we anticipate a very high variance of roughly 180 degrees in the heading and a variance of 3 m/s in the walking speed of the pedestrian. These assumptions lead us to come up with three possible proposal sizes for the sliding window that will generate the training proposals for our method. The varying size of training proposals or sub-images give our model the ability to *"adapt"* and provide superior accuracy once CSI observations related to immediate and extended neighborhood are available.

b. The following are the driving factors behind each of the proposal sizes.

- $1 \times 1$ *size:* This is essentially a single pixel associated to a single location on the map. This training patch or proposal will likely make the classification suffer from highest level of localization inaccuracy since this contains minimal amount of information.
- $2 \times 2$ *size:* This size of sliding window extends to the immediate neighborhood of the walking pedestrian. For the test observation, the sliding proposal for this size, can only be generated when we have enough CSI datapoints available in the vicinity of the pedestrian. Since 4x more information is present in this proposal, the localization performance improves accordingly.
- $3 \times 3$ *size:* This size of sliding window extends beyond immediate neighborhood of pedestrian

location and goes further out into the map-grid. This proposal incorporates 9x more information in the proposal compared to a single cell-based proposal, thus intuitively giving us the maximum space diversity.

c. After extracting proposals of varying sizes, we warp the proposed patches or sub-images into a single size e.g. 3x3x270. These warped proposals are used to train a hybrid CNN-LSTM model as shown in figure 8, for location classification.

d. The classification output is generated as a 37x1 confidence vector ($S_c$). Here each row corresponds to a location in the training dataset (including the null location or no detected location output). The value of each element represents the probability of the classified pedestrian to be present at the corresponding location.

The hybrid CNN-LSTM model which is trained for the generated proposals in step c, is explored in detail in subsection 4.2.3.

### 4.2.3. Hybrid CNN-LSTM based Learning Model

After carefully trying multiple hyper-parameters, we ended up achieving optimal location classification performance via using the hybrid architecture shown in figure 8. Here parameters $U$ and $V$ represent the maximum *Information Adaptive Sliding Window Size* used to extract the proposals from the CSI Fingerprint Map. We can make following observations that sum up our findings at the end of our model development effort. These statements also indicate the suitability of hybrid CNN-LSTM model for our problem.

a. Subject's path trajectory in space is highly variant and as such, no long-term dependency patterns were expected to exist. This was indicated to us when we increased the recurrent weight for the *L2 Regularization Factor* for Forget gate to 4X its default value of 0.001. This penalized our model for forgetting and in effect made our model remember much longer temporal patterns. Under this scenario the model accuracy dropped. Bringing this value to 2X provided us with optimal results. This means there exist some short-term temporal patterns that brought up model accuracy for 2X weight factor.

b. Increasing the *number of CNN layers* adversely affected the overall accuracy of our method. This was primarily caused due to the unsuitable level of learned spatial features provided to LSTM.

c. Increasing the *number of hidden units* in LSTM Layers caused the overfitting to occur. This indicates that having too many parameters available for the model to learn can easily make the model memorize the training dataset.

d. Dropout Layers were not introduced since overfitting was largely addressed via adjusting *L2 Regularization Factor* in LSTM layers.

### 4.2.4. Particle Filtering approach for Trajectory Hypothesis Selection

We propose a particle filtering approach that removes location misclassifications generated by Hybrid CNN-LSTM based learning model. The particle filtering approach consists of two phases namely Multi-Trajectory Hypothesis Generation and Particle Filter based Hypothesis Selection.

*Multi-Trajectory Hypothesis Generation:* The goal of this module is to generate multiple trajectory beliefs based on location confidence values $S_c^t$, evaluated by CNN-LSTM at time *t*. This module requires two parameters to generate trajectory beliefs. The first parameter *f* represents the number of weighted random samples that are chosen from the location belief vector $S_c$. At each time $t$, an output is generated by CNN-LSTM model. Then $f$ samples, weighted by the corresponding probabilities contained in the vector $S_c^t = [c_1^t, c_2^t, \ldots c_{37}^t]$, are chosen. This process is repeated for $n$ timesteps where $n \geq 3$. Here, $n$ is our second parameter for this module. The intuition behind its value is that the trajectory length sent to a particle filter must contain at least 3 observations i.e. we expect the pedestrian to have at least travelled 3 meters before we can analyze the trajectory for conformance with human walk behavior (provided we assume that grid-cell size is 1 sq. meter). At the end of $n$ sampling attempts, we have $n$ vectors of length $f$ each. Now an exhaustive list of possible trajectories $H = \{H_1, H_2, \ldots, H_k\}$, each consisting of $n$ locations is generated, Thus, the total number of generated trajectories is $k = n \times f$. Since the trajectory datapoints points are at least 1 meter apart and are generated approximately every 2 seconds, this does not set well with our particle filter that needs location updates at least every second to converge at a location and heading estimate. To address this, we up-sample the trajectories via interpolation so that the number of updates matches the particle filter requirement. Now each hypothesis $H_b$ has $2k$ location updates for our particle filter. $b$ here is the trajectory hypothesis index.

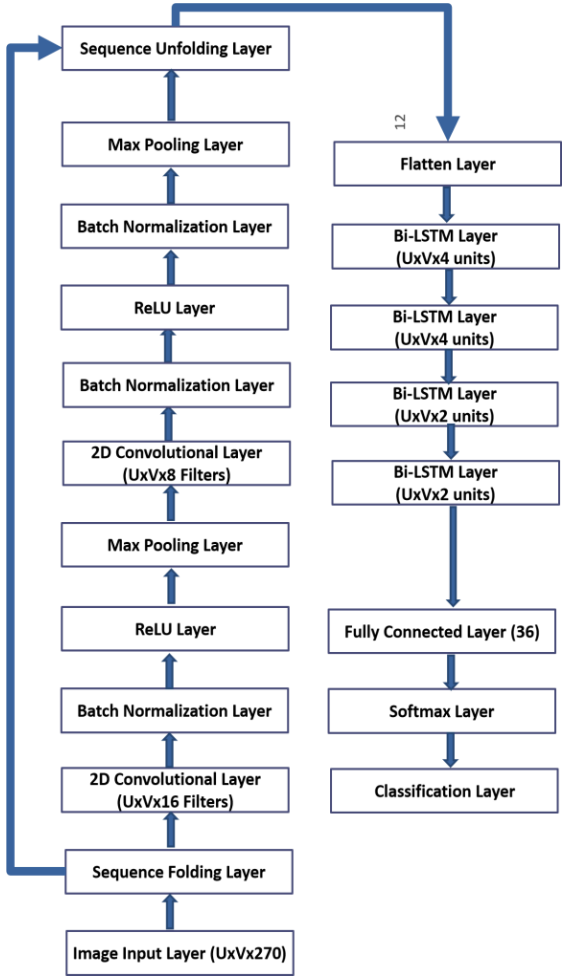

Figure 8. Hybrid CNN-LSTM Learning Model

*Hypothesis Selection via Particle Filter:* The idea behind

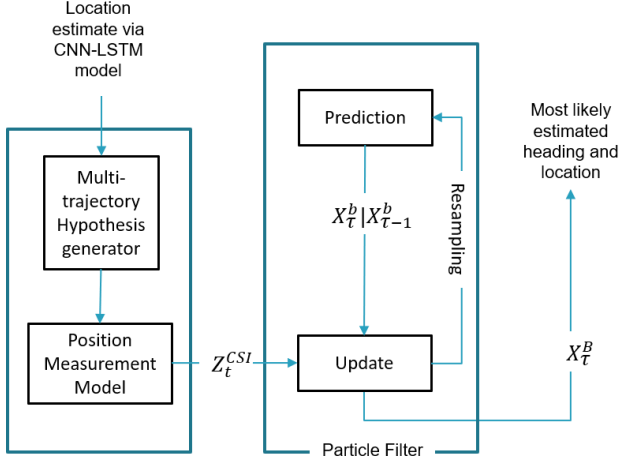

Figure 9. Flowchart outlining Particle Filter based Hypothesis Selection module inputs and outputs and its integration with the overall system

this particle filter-based hypothesis approach is that it estimates location and heading of multiple hypothetical pedestrians while only using the change in pedestrian location over time. The motion model of a pedestrian proposed in [20], keeps multiple records of possible feet locations for a pedestrian hypothesis and verifies, based on a new location update, whether the hypothesis conforms to a realistic walk model or not. We need to track the feet positions so that we are also able to keep track of the stride as it is the distance between two feet of pedestrian. In case we do not use feet position-based walk model, we would have required step size estimation from a dedicated sensor. The hypothesis selection module is our implementation of the above stated approach. It accepts multiple hypothesis as input and initializes a separate set of particles for each of the provided hypothesis $H_b$. The output of the hypothesis selection module consists of a most likely path trajectory $Tr_B$ followed by the pedestrian. It must be noted here that $Tr_B$ is a fine-grained trajectory as particle filter estimates the pedestrian location every 200 milliseconds. Notice that $B$ is the index of the selected trajectory among the given set of hypotheses. A flowchart highlighting inputs and outputs of *Hypothesis Selection Module* and its integration with the overall system is shown in figure 9. The following steps outline the overall process of hypothesis selection.

a. A state vector $X_\tau$ defines the parameters that are estimated for each pedestrian hypothesis. Each particle belonging to each pedestrian hypothesis $H_b$, is initialized through prior belief values to its corresponding state vector $X_\tau^{b,j}$, where $i$ is the hypothesis index, $j$ represents the particle index and $\tau$ represents the particle filter estimation time index which has a higher frequency compared to CNN-LSTM output time index $t$.

$$X_\tau^{b,j} = \{l_{x_\tau}^{b,j}, l_{y_\tau}^{b,j}, r_{x_\tau}^{b,j}, r_{y_\tau}^{b,j}, \theta_\tau^{b,j}, \gamma_\tau^{b,j}, S_\tau^{b,j}, T_\tau^{b,j}\} \quad (4)$$

Here, the first four terms are the positions for left and right feet of pedestrian respectively, while last four terms represent the heading, walk phase, stride and step-period of the pedestrian. All values estimated with respect to the CSI Fingerprint Map reference frame. Hence multi-hypothesis particle filter estimates the location, heading and other relevant parameters for each pedestrian hypothesis via evaluating the probability $P[X_\tau^b | Z_{0:t}^{b,CSI}]$. Here $Z_{0:t}^{b,CSI}$ are all past location updates for a certain hypothesis $H_b$ at time $t$. Each particle is associated with a weight $w_\tau^{b,j}$ which specifies the extent of the particle contribution to the underlying probability density of the pedestrian's state $P[X_\tau^b | Z_{0:t}^{b,CSI}]$.

b. The prediction step for the particle filter is outlined in detail in [20]. In this work we would like to comment on the significance of the update step and the measurement model used for updating the particle filter state. For an incoming location update $Z_t^{b,CSI} = (x_t^{b,i}, y_t^{b,i}, c_t^{b,i})$ at time $t$, for a pedestrian hypothesis $b$, we evaluate the following factors that determine the contribution of this update towards an estimate convergence.

$$p_L = \frac{1}{\sqrt{2\pi}(1-c_t^{b,i})} \sum_j e^{-\frac{\left(l_{x_\tau}^{b,j} - x_t^{b,i}\right)^2 + \left(l_{y_\tau}^{b,j} - y_t^{b,i}\right)^2}{2(1-c_t^{b,i})^2}}$$

$$p_R = \frac{1}{\sqrt{2\pi}(1-c_t^{b,i})} \sum_j e^{-\frac{\left(r_{x_\tau}^{b,j} - x_t^{b,i}\right)^2 + \left(r_{y_\tau}^{b,j} - y_t^{b,i}\right)^2}{2(1-c_t^{b,i})^2}}$$

$$d0 = (r_{x_\tau}^{b,j} - l_{x_\tau}^{b,j})\cos\theta_\tau^{b,j} + (r_{x_\tau}^{b,j} - l_{x_\tau}^{b,j})\sin\theta_\tau^{b,j}$$

$$r0 = -S_\tau^{b,j}\cos(\gamma_\tau^{b,j})$$

$$p_B = \frac{1}{\sqrt{2\pi}h} e^{-\frac{(d0-r0)^2}{2h^2}}$$

$$P(w_\tau^{b,j}|Z_t^{b,CSI}) = p_L \cdot p_R \cdot p_B$$

$$w_\tau^{b,j} = \frac{P(w_\tau^{b,j}|Z_t^{b,CSI})}{\sum_j P(w_\tau^{b,j}|Z_t^{b,CSI})}$$

Factors $p_L$ and $p_R$ contribute towards a particle weight which help the particle draw closer to where $Z_t^{b,CSI}$ believes the pedestrian is at. Factor $p_B$ ensures via contribution towards particle weight that the pedestrian stride and heading is more consistent with the walk model presented in [20]. $d0$ is the distance between the right foot and the left foot of the pedestrian while $r0$ can be described as the reference walking pattern given in [20].

c. Particle filter treats each location within each hypothesis $H_b$ as an update to a tracked pedestrian state $X_\tau^b$. After all updates from all hypothetical trajectories are processed, pedestrian hypothesis with low-confidence values will tend to have high variance in their gaussian distribution. Thus, our measurement model presented in the last step will ensure that a low probability for particle sampling is generated for such a pedestrian hypothesis. This will make the particle filter track the poor hypothesis only and only in the case where such a hypothesis follows a human walking motion model in a strict sense. Such an approach will tend to reduce the sum of all particle weights to zero quickly i.e. $\sum_{j=1}^{J} w_\tau^{b,j} \to 0$. Such a condition requires resampling but having to resample more frequently will trigger our hypothesis tracker to discard a hypothesis that encounters frequent resampling. This process leaves us with at least one hypothesis trajectory $Tr_B$ that best conforms the constraints of human walk motion model.

d. In case a convergence is achieved for more than one trajectory, we evaluate the average confidence value for each competing trajectory via taking the average of corresponding confidence value vector $S_c^t$. The trajectory having the highest average confidence value is termed as the most likely trajectory $Tr_B$.

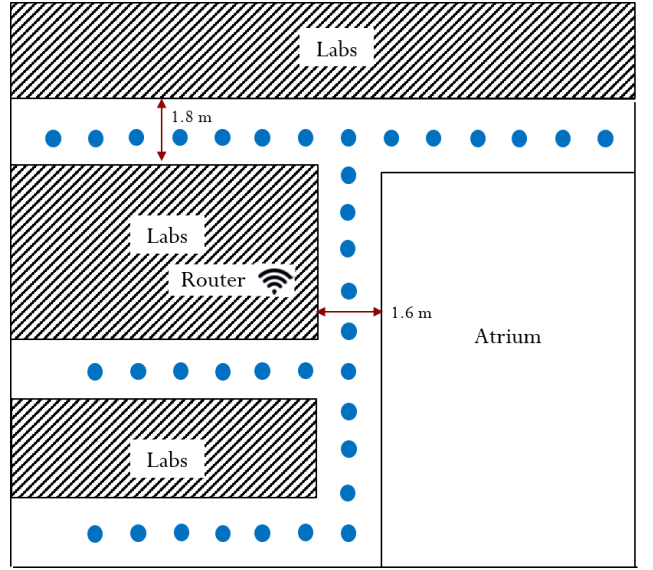

Figure 10. 10×14m Experiment Area. Blue dots indicate reference points for collecting training observations.

e. Lastly, after a likely trajectory is established, the trajectory is mapped onto the grid-cells defined during the training dataset collection phase based on equidistant reference locations. This allows us to evaluate accuracy improvement due to Particle Filter-based Hypothesis Selection Module (PF-HSM).

## 5. Results and Performance Evaluation

### 5.1. Experiment Setup

We use a low-power pocket held embedded PC as a Wi-Fi device for collecting test data set. This device is installed with a power optimized Ubuntu LTS configuration; thus, the device can last upwards of 9 hours of data collection activity. The device has an Intel Wi-Fi Link 5300 NIC wireless LAN card, along with a 3 Port MIMO Antenna. Hybrid CNN-LSTM model training was performed suing a High-Performance Computing node installed with an NVIDIA K80 Accelerator and 20 GB of RAM. We verify our model in an indoor corridor scenario as shown in figure 10. The whole experiment area is about 14 m by 10 m. It must be mentioned here that there exists no line of sight (LOS) between the router and the reference points for training observation collection.

### 5.2. Methodology

While the reference locations are kept equidistant (1 meter apart) and fixed, test locations have not such restrictions placed on these. A subject is allowed to walk through the corridors and a stream of observations is collected with each observation 200 milliseconds apart in time. A consecutive subset of these observations is extracted as a trajectory and tested on our proposed method. There is no restriction on the temporal distance between two test observations in the sequence. Test location sequences that last between 6 seconds to 10 seconds, are extracted from the test CSI data stream and fed into the proposed system for localization and tracking.

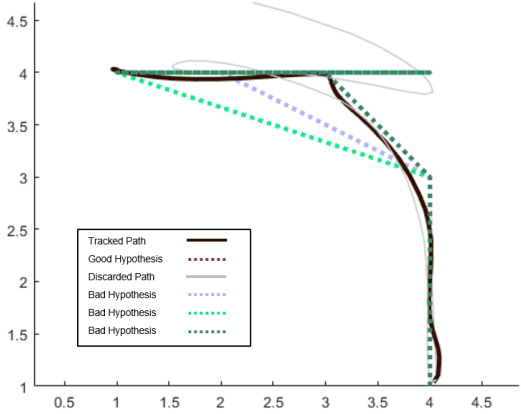

Figure 11. Among several trajectory hypothesis, all but one is rejected due to non-conformance to human walk-model by Particle Filter HSM

### 5.3. Method Comparison

The first method that we shortlisted for accuracy comparison with the proposed method is called DeepFi [15]. This method has been widely employed for comparisons for CSI based positioning method and can provide a good baseline for achievable localization accuracy via CSI based fingerprinting. Moreover, this method also uses a Bayes probability model based post-processing technique to estimate final positioning of the subject. Since we also employ a particle filter-based location estimation technique, DeepFi stands out to be a good comparison candidate.

The second method chosen for comparison is known as ConFi[17]. As already discussed earlier, this method also uses 2D+channel CNN for location classification, thus we wanted to have a closely related and relatively recent technique present in the comparison table as well.

The last method for comparison was developed by us. This method is an LSTM based classifier that only uses temporal features of the CSI fingerprints and completely ignores the spatial relationships between the fingerprints.

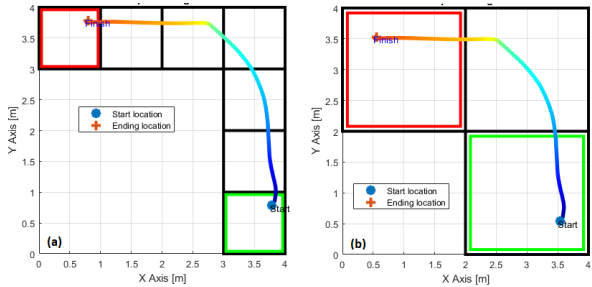

Figure 12. Trajectory to grid-cell mapping to evaluate accuracy improvement after PF-HSM application

Having such a method in the comparison mix, highlights the significant improvement in accuracy, due to the integration of the spatial feature into the learning model. This model uses the fingerprint format ($FP_{x,y}^t$) given by equation 2. The comparison of distance error between LSTM, DeepFi, ConFi and proposed method is represented as a Cumulative Density Function (CDF) chart in figure 13. The test dataset used to evaluate this comparison was collected in a moderately dynamic environment i.e. during lunch hour at 1pm on a weekday. The mean errors all compared methods is presented in Table 1.

Table 2: Mean Errors Comparison

| Method | Mean Error | Std. dev |
|---|---|---|
| | Value in meters | |
| LSTM | 2.20 | 1.80 |
| DeepFi | 1.99 | 1.74 |
| ConFi | 1.43 | 1.15 |
| CNN-LSTM | 0.79 | 0.68 |

*Note: The mean error reported for CNN-LSTM is for the system output before the application of particle filter-based hypothesis selection method*

### 5.4. Effect of Parameters and Hypothesis Selection Method on Localization Performance

Parameters that have a significant impact on localization performance of the method include grid-size for training dataset collection, Information Adaptive Sliding Window Size, Pedestrian walking speed. Apart from these parameters, Hypothesis Selection Method, reduces the Mean Distance Error to as low as 0.07 m for a specific combination of parameters. Table 2 lists the corresponding localization errors before and after the application of HSM for several parameter combinations.

It is evident from the experimental data that learning spatial features alongside the temporal features on a map-based representation has a significantly positive impact of localization accuracy. PF-HSM application over the CNN-LSTM model output provides us with an average 38% improvement in terms of distance error measure. The primary reason behind such a boost in accuracy is the fact that particle filter is easily able to reject mis-classified locations that lie farther apart in terms of distance, the human walk motion model is easily able to filter out pedestrian hypothesis that either try to breach human walking speeds or corresponding limits on turning angles. An instance of such hypothesis rejection from the real test dataset is shown in figure 11. The selected trajectory is then mapped onto the grid-cells defined during the training dataset collection phase. The grid-cells that overlap selected trajectory points are then used to evaluate the improved localization accuracy listed in the last column of table 2. An instance of such a mapping is shown in figure 12. Due to the non-availability of accurate sensors that could measure accurate pedestrian heading for each location at a resolution of 5Hz or below, a simulated human walk motion model [21] was implemented to generate human-like motion trajectories. This motion model is easily configurable, and bounds can be set on normally distributed pedestrian walk phase, speed, stride and heading direction. The simulated noisy data along with the ground truth was fed to PF-HSM to assess the tracking performance of particle filter. The accuracy results for tracked location, heading and stride length are presented in the ground truth vs. estimation chart presented in figure 14.

## 6. Conclusion

The proposed method delivers robust results in terms of pedestrian localization and heading. A temporally stable and diverse sanitized CSI phase value signal is used for fingerprinting. The novelty of the proposed method lies in representation of the signal to 2D mapping domain and then exploiting this representation to be used in location classification using a hybrid CNN-LSTM learning model. Not only is the proposed method more accurate than contemporary methods in terms of pedestrian localization, it also manages to deliver estimated pedestrian heading without the use of any additional sensors such as an IMU or Magnetometer. The accuracy and stability of generated trajectory and heading based on Wi-Fi signals are verified via experimental results.

Table 3

| Grid-size for Training Dataset | Information Adaptive Sliding Window Size | Walking Speed | Mean Distance Error before PF-HSM | Mean Distance Error after PF-HSM |
|---|---|---|---|---|
| in meters | in pixels | m/s | in meters | in meters |
| 1 | 1×1 | 1 | 1.31 | 0.67 |
| 1 | 1×1 | 3 | 1.93 | 0.75 |
| 1 | 2×2 | 1 | 0.75 | 0.62 |
| 1 | 2×2 | 3 | 0.88 | 0.69 |
| 1 | 3×3 | 1 | 0.31 | 0.21 |
| 1 | 3×3 | 3 | 0.32 | 0.22 |
| 2 | 1×1 | 1 | 1.55 | 0.58 |
| 2 | 1×1 | 3 | 1.82 | 0.64 |
| 2 | 2×2 | 1 | 0.61 | 0.49 |
| 2 | 2×2 | 3 | 0.89 | 0.50 |
| 2 | 3×3 | 1 | 0.12 | 0.07 |
| 2 | 3×3 | 3 | 0.12 | 0.07 |

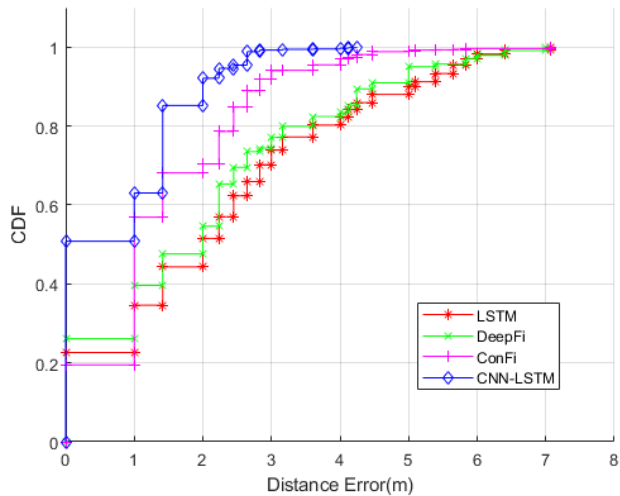

Figure 13. Pedestrian Positioning Error Performance Comparison

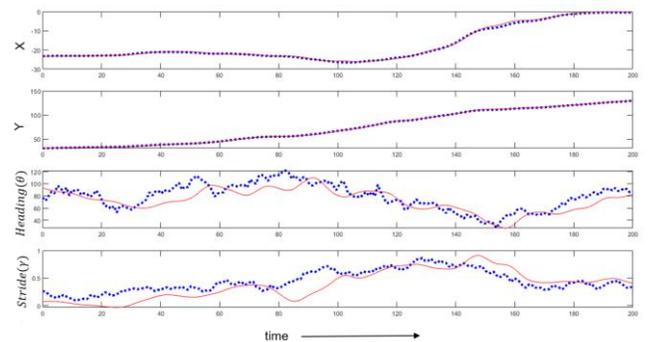

Figure 14. Particle Filter Tracking Accuracy: Simulation results for Tracked vs. Ground truth state differences.

## 7. References

[1] X. Wang, Z. Yu, and S. Mao, "DeepML: Deep LSTM for indoor localization with smartphone magnetic and light sensors," in Proc. IEEE ICC 2017, Kansas City, MO, May 2018.

[2] X. Wang, L. Gao, S. Mao and S. Pandey, "CSI-Based Fingerprinting for Indoor Localization: A Deep Learning Approach," in IEEE Transactions on Vehicular Technology, vol. 66, no. 1, pp. 763-776, Jan. 2017.

[3] J. Chung, C. Gulcehre, K. Cho, and Y. Bengio. Empirical evaluation of gated recurrent neural networks on sequence modeling. arXiv:1412.3555, 2014.

[4] X. Wang, L. Gao and S. Mao, "BiLoc: Bi-Modal Deep Learning for Indoor Localization with Commodity 5GHz WiFi," in IEEE Access, vol. 5, pp. 4209-4220, 2017.

[5] J. Torres-Sospedra and A. Moreira, "Analysis of Sources of Large Positioning Errors in Deterministic Fingerprinting," Sensors, vol. 17, no. 12, p. 2736, 2017

[6] Yao, S., Hu, S., Zhao, Y., Zhang, A., & Abdelzaher, T. (2017). DeepSense. Proceedings of the 26th International Conference on World Wide Web - WWW 17. doi:10.1145/3038912.3052577

[7] Wang, X., Yu, Z., & Mao, S. (2018). DeepML: Deep LSTM for Indoor Localization with Smartphone Magnetic and Light Sensors. 2018 IEEE International Conference on Communications (ICC).

[8] Kang J, Lee J, Eom DS. Smartphone-Based Traveled Distance Estimation Using Individual Walking Patterns for Indoor Localization. Sensors (Basel). 2018;18(9):3149. Published 2018 Sep 18. (As an example of GRU implementation) doi:10.1109/icc.2018.8422562

[9] Lohan, Elena Simona, Torres-Sospedra, Joaquín, Leppäkoski, Helena, Richter, Philipp, Peng, Zhe, Huerta, Joaquín "Wi-Fi Crowdsourced Fingerprinting Dataset for Indoor Positioning", MDPI Data journal

[10] Weblink: http://indoorlocplatform.uji.es/databases/all/

[11] Adib, F.; Kabelac, Z.; Katabi, D. Multi-person localization via RF body reflections. In Proceedings of the 12th USENIX Conference on Networked Systems Design and Implementation, Oakland USA, pp.279-292, May 2015.

[12] Colone, F.; Woodbridge, K.; Guo, H.; Mason, D.; Baker, C.J. Ambiguity function analysis of wireless LAN transmissions for passive radar. IEEE Trans. Aerosp. Electron. Syst, 2011, vol. 47, no. 1, pp. 240-264.

[13] Khalili, A.; Soliman, A. A. Track before mitigate: aspect dependence-based tracking method for multipath mitigation. Electronics Letters, 2016, vol. 52, no. 4, pp. 316-317.

[14] Khalili, A.; Soliman, A. A.; Asaduzzaman, M. Quantum particle filter: a multiple mode method for low delay abrupt pedestrian motion tracking. Electronics Letters, 2015, vol. 51, no. 16, pp. 1251-1253.

[15] X. Wang, L. Gao, S. Mao and S. Pandey, "DeepFi: Deep learning for indoor fingerprinting using channel state information," 2015 IEEE Wireless Communications and Networking Conference (WCNC), New Orleans, LA, 2015, pp. 1666-1671.

[16] Wang, Xuyu et al. "CiFi: Deep convolutional neural networks for indoor localization with 5 GHz Wi-Fi." 2017 IEEE International Conference on Communications (ICC) (2017): 1-6.

[17] H. Chen, Y. Zhang, W. Li, X. Tao and P. Zhang, "ConFi: Convolutional Neural Networks Based Indoor Wi-Fi Localization Using Channel State Information," in IEEE Access, vol. 5, pp. 18066-18074, 2017.

[18] Qian K, Wu C, Yang Z, et al. PADS: Passive Detection of Moving Targets with Dynamic Speed using PHY Layer Information. In: Proc. IEEE ICPADS'14. Hsinchu, Taiwan; 2014. p. 1–8.

[19] S. Riyaz, K. Sankhe, S. Ioannidis and K. Chowdhury, "Deep Learning Convolutional Neural Networks for Radio Identification," in IEEE Communications Magazine, vol. 56, no. 9, pp. 146-152, Sept. 2018.

[20] M. Emaduddin and D. A. Shell, "Estimation of pedestrian distribution in indoor environments using multiple pedestrian tracking," in IEEE Intl. conf. on Robotics and Automation, 2009.

[21] Straub, J. (2010). Pedestrian Indoor Localization and Tracking using a Particle Filter combined with a learning Accessibility Map (Dissertation). Technische Universität München, Munich, Germany.